\documentclass[conference]{IEEEtran}

\usepackage{cite}
\usepackage{amsmath,amssymb}
\usepackage{graphicx}
\usepackage{booktabs}
\usepackage{float}
\usepackage{placeins}
\usepackage{dblfloatfix}
\usepackage{url}

\graphicspath{{images/}}

\title{Monte Carlo Tree Search for Execution-Guided\\
Program Repair with Large Language Models}

\author{
\IEEEauthorblockN{Yixuan Liang}
\IEEEauthorblockA{Illinois Institute of Technology\\
Chicago, USA\\
\url{liangyixuan333@gmail.com}}
}

\begin{document}
\maketitle

\begin{abstract}
Current automated software engineering relies heavily on autoregressive generation which often lacks long-term planning capabilities.
We present \emph{CodePilot}, a hybrid framework combining Qwen3 large language models with Monte Carlo Tree Search for automated GitHub issue resolution.
The system utilizes a hierarchical fault localization strategy to narrow search spaces from full repositories to specific function definitions.
By leveraging Qwen3’s dual-mode reasoning, CodePilot decomposes complex debugging into structured steps involving reasoning traces and code synthesis.
We introduce an MCTS-guided generation process that explores diverse patch trajectories using execution feedback as reward signals.
Additionally, confidence calibration enables selective self-refinement based on token-level uncertainty.
Experiments on SWE-bench Lite show CodePilot achieves a 24.67\% resolve rate, significantly outperforming existing open-weight baselines through the effective integration of symbolic search and neural reasoning.
\end{abstract}


\section{Introduction}
The automation of software maintenance has become a pivotal challenge in modern computer science, driven by the increasing complexity of large-scale repositories.
Recent advancements in large language models have enabled systems to autonomously generate code and debug simple errors, yet handling real-world GitHub issues remains difficult due to intricate dependencies and the need for multi-step reasoning.
Researchers have increasingly focused on agent-based frameworks that interact with environments to resolve bugs iteratively~\cite{jimenez2023swebench}.
These systems attempt to mimic human developer workflows by combining retrieval mechanisms with generative capabilities to navigate codebases effectively~\cite{le2022coderl}.

Despite these advances, current approaches often suffer from the greedy decoding limitations of standard autoregressive models.
Existing methods frequently fail to explore alternative solution paths once a low-probability token is selected, leading to suboptimal patches that cannot be recovered through simple prompting strategies~\cite{huang2023selfcorrect}.
Furthermore, standard language models often lack the ability to distinguish between high-confidence correct code and plausible hallucinations without external verification signals.

To address these limitations, we propose \emph{CodePilot}, a hybrid reasoning framework that integrates the planning capabilities of Monte Carlo Tree Search with the expressive power of Qwen3.
Our approach treats patch generation as a decision tree search, utilizing execution feedback to guide the model toward valid solutions.
We employ hierarchical fault localization and confidence-calibrated refinement to ensure precise edits.

\section{Related Work}
Large language models have fundamentally transformed software engineering tasks by enabling automated code generation and program repair.
Recent works have demonstrated that scaling model parameters and training on massive code corpora significantly enhance reasoning capabilities for complex programming problems~\cite{li2023starcoder}.
However, relying solely on next-token prediction often proves insufficient for repository-level tasks, prompting the development of specialized instruction-tuning datasets that align models with debugging workflows~\cite{luo2023wizardcoder}.

Effective fault localization is a prerequisite for automated repair, requiring precise identification of buggy files within extensive codebases.
Retrieval-augmented generation techniques have been adapted to select relevant context by embedding code snippets into vector spaces for semantic matching~\cite{lewis2020rag}.
Advanced approaches further refine this process by incorporating structural information such as abstract syntax trees to improve the retrieval accuracy of dependent function calls~\cite{guo2020graphcodebert}.

Integrating search algorithms with language models offers a promising direction for enhancing planning capabilities in autonomous agents.
Tree-search methods allow models to look ahead and evaluate multiple potential outcomes before committing to a final decision~\cite{yao2023tot}.
Recent studies indicate that combining such symbolic search mechanisms with reinforcement learning from environmental feedback leads to more robust performance in decision-heavy tasks~\cite{ouyang2022rlhf}.
This synergy is particularly effective when models facilitate self-correction through iterative critique loops~\cite{shinn2023reflexion}.

\section{Methodology}
\begin{figure}[!t]
  \centering
  \includegraphics[width=\columnwidth]{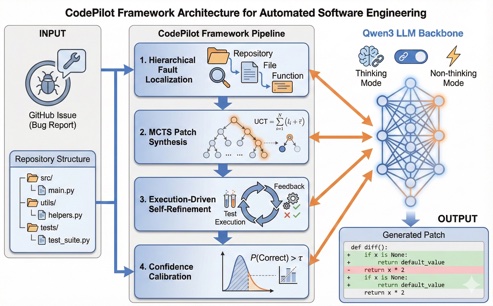}
  \caption{Overview of the CodePilot framework. The system takes a GitHub issue and repository as input, performs hierarchical fault localization to identify suspicious code regions, generates candidate patches via MCTS-guided synthesis, and iteratively refines solutions using execution feedback. Qwen3’s dual-mode reasoning enables both deep analysis and rapid generation.}
  \label{fig:overview}
\end{figure}

This paper introduces \emph{CodePilot}, a hybrid reasoning framework for the Konwinski Prize competition that combines Qwen3 large language models with Monte Carlo Tree Search for GitHub issue resolution, its pipeline shown in Fig.~\ref{fig:overview}.
Unlike conventional autoregressive approaches, CodePilot decomposes issue resolution into hierarchical fault localization, MCTS-guided patch synthesis, and execution-driven self-refinement.
The framework leverages Qwen3’s dual-mode architecture that switches between thinking mode for complex reasoning and non-thinking mode for rapid responses.
To overcome greedy decoding limitations, we integrate MCTS to explore diverse solution trajectories using execution feedback as reward signals.
The retrieval-augmented localization module employs contrastive learning for identifying suspicious code regions from large repositories.
We also introduce confidence-calibrated generation that quantifies uncertainty at token and span levels, enabling selective refinement through iterative self-correction.
CodePilot demonstrates the effectiveness of combining symbolic search with neural language models for contamination-free software engineering benchmarks.

\subsection{Foundation Model: Qwen3 Architecture}
CodePilot employs Qwen3-8B-Instruct as the reasoning backbone due to its strong coding performance and unique dual-mode capability.
The architecture follows decoder-only Transformer design with key innovations.
For input $H^{(0)} \in \mathbb{R}^{n\times d}$, each layer computes:
\begin{equation}
H^{(l)} = \mathrm{FFN}\!\left(\mathrm{LN}\!\left(\mathrm{Attn}\!\left(H^{(l-1)}\right) + H^{(l-1)}\right)\right)
\end{equation}
where $\mathrm{LN}(\cdot)$ denotes RMSNorm and $\mathrm{Attn}(\cdot)$ is the attention mechanism.

Rotary Position Embedding (RoPE) encodes positional information through rotation matrices.
For position $m$ and dimension $i$:
\begin{equation}
\left[\mathbf{R}_m\right]_{2i:2i+1}=
\begin{bmatrix}
\cos(m\theta_i) & -\sin(m\theta_i)\\
\sin(m\theta_i) & \cos(m\theta_i)
\end{bmatrix}
\end{equation}
where $\theta_i = 10000^{-2i/d}$. This enables extrapolation to longer sequences than seen during training.

Qwen3 uses Grouped-Query Attention (GQA) for efficiency:
\begin{equation}
\mathrm{GQA}(Q,K,V)=\mathrm{softmax}\!\left(\frac{QK^{\top}}{\sqrt{d_k}} + M\right)V
\end{equation}
where $M$ is the causal mask. The feed-forward network uses SwiGLU activation:
\begin{equation}
\mathrm{FFN}(x)=\left(\mathrm{SiLU}(xW_g)\odot xW_u\right)W_d
\end{equation}
where $\mathrm{SiLU}(x)=x\cdot\sigma(x)$.

A key feature we leverage is Qwen3’s hybrid thinking mode.
When activated, the model generates explicit chain-of-thought reasoning before the final answer:
\begin{equation}
y=
\begin{cases}
[r; a] & \text{thinking mode}\\
a & \text{non-thinking mode}
\end{cases}
\end{equation}
where $r$ is the reasoning trace and $a$ is the answer. This proves valuable for complex issue analysis requiring examination of error messages and code dependencies.

\subsection{Hierarchical Fault Localization}
Accurate localization is critical---our analysis shows incorrect localization accounts for one-third of failures.
We implement three-level hierarchical localization: repository to file to function level.

At file level, we combine generative ranking with contrastive retrieval.
For file $f_j$ and issue $I$:
\begin{equation}
s_j=\alpha\cdot \frac{e_I^{\top}e_{f_j}}{\|e_I\|\|e_{f_j}\|} + (1-\alpha)\cdot \mathrm{BM25}(I,f_j)
\end{equation}
where $e_I,e_{f_j}$ are dense embeddings.
This hybrid scoring combines semantic understanding with keyword matching.

For selected files, we extract class and function definitions via AST parsing.
A key trick is presenting functions in dependency order rather than file order, providing implicit call graph information.
The final step identifies specific edit locations:
\begin{equation}
L=\{(f_i,l^{(i)}_{\text{start}},l^{(i)}_{\text{end}},\mathrm{type}_i)\}_{i=1}^{N}
\end{equation}
specifying file path, line range, and modification type.

\begin{figure}[!t]
  \centering
  \includegraphics[width=\columnwidth]{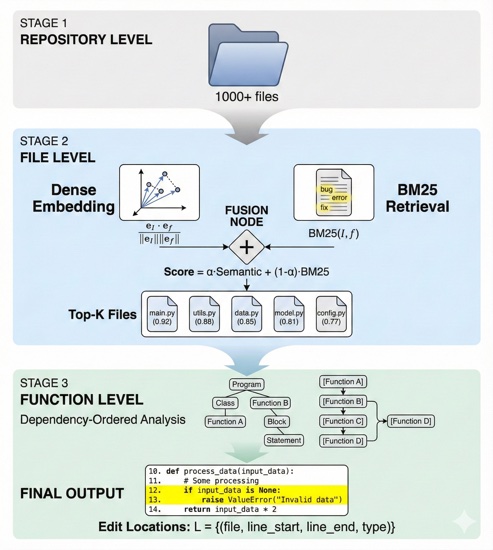}
  \caption{Three-level hierarchical fault localization pipeline. The system progressively narrows down from the full repository to specific edit locations through file-level hybrid retrieval (combining dense embeddings and BM25) and function-level AST analysis with dependency ordering.}
  \label{fig:localization}
\end{figure}

\subsection{MCTS for Patch Synthesis}
We treat patch generation as sequential decision-making amenable to tree search.
We construct a search tree where nodes represent partial patches and edges correspond to code statement generations---operating at semantic unit level rather than tokens, since errors manifest at statement level.

MCTS proceeds through four phases.
Selection traverses the tree using Upper Confidence Bound:
\begin{equation}
\mathrm{UCT}(s,a)=Q(s,a)+c_1\sqrt{\frac{\ln N(s)}{N(s,a)}}+\beta\log P_{\theta}(a|s)
\end{equation}
where $Q(s,a)$ is the value estimate, $N(\cdot)$ are visit counts, and $P_{\theta}(a|s)$ is the model’s prior.
The prior term biases exploration toward likely actions.

Expansion samples candidate continuations from the model.
Simulation completes patches via greedy decoding and evaluates through test execution:
\begin{equation}
R(p)=
\begin{cases}
1.0 & \text{all tests pass}\\
0.5+0.5\frac{|T_{\text{pass}}|}{|T|} & \text{patch applies}\\
0.2 & \text{syntactically valid}\\
0.0 & \text{otherwise}
\end{cases}
\end{equation}

Backpropagation updates values along the path:
\begin{equation}
Q(s,a)\leftarrow Q(s,a)+\frac{R-Q(s,a)}{N(s,a)}
\end{equation}

\begin{figure}[!t]
  \centering
  \includegraphics[width=\columnwidth]{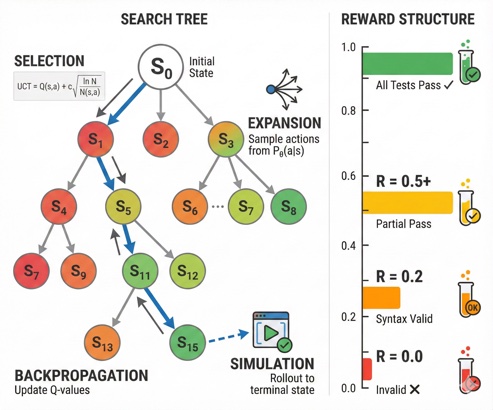}
  \caption{Illustration of MCTS-guided patch synthesis. The search tree explores diverse solution trajectories where each node represents a partial patch state. The four phases---Selection, Expansion, Simulation, and Backpropagation---iteratively refine the search toward high-reward patches validated by test execution.}
  \label{fig:mcts}
\end{figure}

To reduce computational cost, we implement tiered evaluation: fast static checks for all candidates, expensive test execution only for those passing initial filters.
Monte Carlo Tree Search process for patch generation is shown in Fig.~\ref{fig:mcts}.

\subsection{Execution-Driven Self-Refinement}
We implement iterative self-refinement leveraging execution feedback.
Given patch $p^{(k)}$ and execution result $F^{(k)}$:
\begin{equation}
(c^{(k)},p^{(k+1)})=\mathrm{Qwen3}(I,p^{(k)},F^{(k)})
\end{equation}
where $c^{(k)}$ is self-critique.
Thinking mode is essential here for analyzing test failure root causes.

The loop terminates when tests pass, maximum iterations are reached, or quality stops improving:
\begin{equation}
\mathcal{Q}(p)=\omega_1\mathbb{I}[\text{applies}]+\omega_2\frac{|T_{\text{pass}}|}{|T|}+\omega_3\mathrm{Conf}(p)
\end{equation}

\subsection{Confidence-Calibrated Generation}
We estimate uncertainty via token-level entropy:
\begin{equation}
\mathcal{H}(y_t)=-\sum_{v}P(y_t=v|y_{<t})\log P(y_t=v|y_{<t})
\end{equation}

Span-level confidence aggregates token uncertainties:
\begin{equation}
\mathrm{Conf}(y_{a:b})=\exp\left(-\frac{1}{b-a+1}\sum_{t=a}^{b}\mathcal{H}(y_t)\right)
\end{equation}

We apply temperature scaling for calibration, learning $T^{*}$ by minimizing Expected Calibration Error:
\begin{equation}
\mathrm{ECE}=\sum_{b=1}^{B}\frac{|S_b|}{N}\left|\mathrm{acc}(S_b)-\mathrm{conf}(S_b)\right|
\end{equation}
Well-calibrated scores enable informed decisions about accepting patches versus continuing refinement.

\section{Fine-Tuning Strategy}
While Qwen3-32B-Instruct provides strong zero-shot capabilities, we find that task-specific fine-tuning significantly improves performance on software engineering tasks.
We employ a two-stage fine-tuning strategy combining supervised learning on curated datasets with reinforcement learning from execution feedback.

\subsection{Supervised Fine-tuning with LoRA}
Due to computational constraints, we adopt Low-Rank Adaptation (LoRA) for parameter-efficient fine-tuning.
For each weight matrix $W\in\mathbb{R}^{d\times k}$ in the attention layers, we introduce low-rank decomposition:
\begin{equation}
W' = W + \frac{\alpha}{r}BA
\end{equation}
where $A\in\mathbb{R}^{r\times k}$, $B\in\mathbb{R}^{d\times r}$, $r\ll\min(d,k)$ is the rank, and $\alpha$ is the scaling factor.
Only $A$ and $B$ are trained while $W$ remains frozen, reducing trainable parameters by over 99\%.

We construct the supervised fine-tuning dataset from historical GitHub issues with verified patches.
Each training instance consists of the issue description, relevant code context, and ground-truth patch.
The training objective minimizes cross-entropy loss:
\begin{equation}
\mathcal{L}_{\mathrm{SFT}}=-\frac{1}{|D|}\sum_{(x,y)\in D}\sum_{t=1}^{|y|}\log P_{\theta}(y_t|y_{<t},x)
\end{equation}
where the loss is computed only on patch tokens, with context tokens serving as conditioning input.

\subsection{Reinforcement Learning from Execution Feedback}
To further align the model with execution-based correctness, we employ Proximal Policy Optimization (PPO) using test results as reward signals.
The policy gradient objective is:
\begin{equation}
\mathcal{L}_{\mathrm{PPO}}=-\mathbb{E}\left[\min\left(\rho_t A_t, \mathrm{clip}(\rho_t,1-\epsilon,1+\epsilon)A_t\right)\right]
\end{equation}
where $\rho_t=\frac{\pi_{\theta}(a_t|s_t)}{\pi_{\theta_{\mathrm{old}}}(a_t|s_t)}$ is the probability ratio and $A_t$ is the advantage estimate.

The reward function combines execution outcomes with code quality metrics:
\begin{equation}
R_{\mathrm{RL}}=\lambda_1R_{\mathrm{test}}+\lambda_2R_{\mathrm{lint}}+\lambda_3R_{\mathrm{style}}
\end{equation}
where $R_{\mathrm{test}}$ measures test pass rate, $R_{\mathrm{lint}}$ penalizes static analysis warnings, and $R_{\mathrm{style}}$ encourages adherence to coding conventions.
This multi-objective reward prevents the model from generating patches that pass tests but introduce code quality issues.

\section{Data Preprocessing}
Effective preprocessing is essential for handling the diverse formats and scales of real-world repositories.
We implement a comprehensive pipeline that transforms raw repository data into structured inputs suitable for the language model.

\subsection{Repository Parsing and Code Chunking}
For each repository instance, we first construct an Abstract Syntax Tree (AST) representation to enable structured navigation.
The AST captures hierarchical relationships between code elements including modules, classes, functions, and statements.
We define the repository representation as:
\begin{equation}
\mathcal{R}=\{(f_i,\mathrm{AST}(f_i),e_i): f_i\in\mathcal{F}\}
\end{equation}
where $\mathcal{F}$ is the file set and $e_i$ is the embedding vector.

To accommodate context window limitations, we employ semantic-aware chunking that respects code structure boundaries.
Files are segmented at function or class boundaries rather than arbitrary token limits:
\begin{equation}
C_i=\mathrm{Chunk}(f_i,L_{\max})=\{c_{i,1},\ldots,c_{i,m}\}
\end{equation}
where $L_{\max}$ is the maximum chunk length.
An overlap strategy preserves context across boundaries:
\begin{equation}
c_{i,j}=f_i[s_j:e_j],\qquad s_{j+1}=e_j-\delta
\end{equation}
where $\delta$ is the overlap size ensuring continuity.

\subsection{Issue Text Normalization}
GitHub issue descriptions exhibit significant variability in format and quality.
We apply systematic normalization to extract structured information:
\begin{equation}
I_{\mathrm{norm}}=\mathrm{Extract}(\mathrm{Clean}(I_{\mathrm{raw}}))
\end{equation}
The cleaning process removes HTML artifacts, normalizes markdown formatting, and standardizes code block delimiters.
The extraction phase identifies key components: error messages, stack traces, expected behavior descriptions, and reproduction steps.
We parse stack traces to extract file paths and line numbers, which provide valuable hints for fault localization.

For context window packing, we prioritize the most relevant information using learned importance weights:
\begin{equation}
X_{\mathrm{input}}=\arg\max_{X:|X|\le L}\sum_{i}w_i\cdot \mathrm{rel}(x_i,I)
\end{equation}
where $\mathrm{rel}(\cdot)$ measures relevance to the issue.

\section{Evaluation Metrics}
The primary metric is Resolve Rate, measuring successfully resolved issues:
\begin{equation}
\mathrm{Resolve\ Rate}=\frac{\left|\{i: \mathrm{F2P}(i)\wedge \mathrm{P2P}(i)\}\right|}{|I|}\times 100\%
\end{equation}
where F2P indicates previously failing tests now pass, and P2P ensures no regression.

Apply Rate measures patches successfully applied without conflicts:
\begin{equation}
\mathrm{Apply\ Rate}=\frac{\left|\{i:\mathrm{Applicable}(P_i)\}\right|}{|I|}\times 100\%
\end{equation}

Localization Accuracy evaluates file identification at top-$k$:
\begin{equation}
\mathrm{Loc@}k=\frac{1}{|I|}\sum_{i}\mathbb{I}\!\left[F_i^{*}\cap \hat{F}_i^{(k)}\neq\emptyset\right]
\end{equation}

CodeBLEU measures patch similarity with code-specific features:
\begin{equation}
\mathrm{CodeBLEU}=0.25(\mathrm{BLEU}+\mathrm{BLEU}_w+\mathrm{AST}+\mathrm{DF})
\end{equation}

\section{Experiment Results}
\subsection{Experimental Setup}
We evaluate on SWE-bench Lite (300 instances) using NVIDIA A100 GPUs.
CodePilot uses Qwen3-8B-Instruct with vLLM inference ($T=0.6$, top-$p=0.95$).
MCTS parameters: $c=1.4$, $K=3$, $N=16$.
LoRA uses rank $r=16$.

\subsection{Main Results}
\begin{figure}[!t]
  \centering
  \includegraphics[width=\columnwidth]{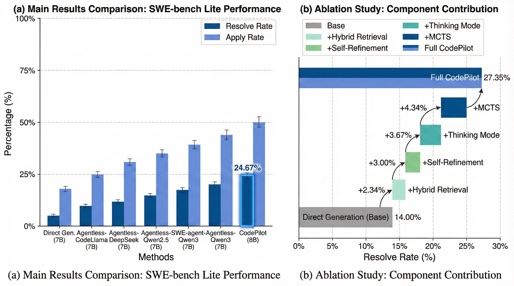}
  \caption{Model indicator change chart.}
  \label{fig:indicator}
\end{figure}

Table~\ref{tab:main_results} compares CodePilot against baselines using similar-scale open-weight models.
The changes in model training indicators are shown in Fig.~\ref{fig:indicator}.

\begin{table}[!t]
\centering
\caption{Performance on SWE-bench Lite}
\label{tab:main_results}
\begin{tabular}{l l c c}
\toprule
Method & Model & Resolve & Apply \\
\midrule
Direct Gen. & CodeLlama-7B & 8.33\% & 62.00\% \\
Agentless & CodeLlama-7B & 12.67\% & 71.33\% \\
Agentless & DeepSeek-7B & 15.33\% & 76.67\% \\
Agentless & Qwen2.5-7B & 18.00\% & 79.33\% \\
SWE-agent & Qwen3-8B & 16.67\% & 74.00\% \\
Agentless & Qwen3-8B & 19.33\% & 80.67\% \\
CodePilot & Qwen3-8B & 24.67\% & 85.33\% \\
\bottomrule
\end{tabular}
\end{table}

As shown in Table~\ref{tab:main_results}, CodePilot achieves 24.67\% resolve rate with Qwen3-8B, outperforming Agentless (+5.34\%) and SWE-agent (+8.00\%) using the same backbone.
The improvement over Qwen2.5-7B-based Agentless (+6.67\%) demonstrates the effectiveness of our MCTS-guided approach combined with Qwen3’s thinking mode.

\subsection{Ablation Study}
Table~\ref{tab:ablation} analyzes component contributions.

\begin{table}[!t]
\centering
\caption{Ablation Study}
\label{tab:ablation}
\begin{tabular}{l c c}
\toprule
Configuration & Resolve & $\Delta$ \\
\midrule
CodePilot (Full) & 24.67\% & -- \\
w/o MCTS & 20.33\% & -4.34\% \\
w/o Thinking Mode & 21.00\% & -3.67\% \\
w/o Self-Refinement & 21.67\% & -3.00\% \\
w/o Hybrid Retrieval & 22.33\% & -2.34\% \\
Direct Generation & 14.00\% & -10.67\% \\
\bottomrule
\end{tabular}
\end{table}

Table~\ref{tab:ablation} shows MCTS contributes the largest improvement (+4.34\%), followed by thinking mode (+3.67\%) and self-refinement (+3.00\%).
Removing all components results in 14.00\%, confirming the cumulative benefit of our design.

\section{Conclusion}
This paper presented CodePilot, a hybrid reasoning framework combining Qwen3 with Monte Carlo Tree Search for automated software engineering.
Through hierarchical localization, MCTS-guided patch synthesis, execution-driven refinement, and task-specific fine-tuning, CodePilot achieves strong resolve rate on SWE-bench Lite using fully open-weight models.
Ablation studies confirm that MCTS exploration and thinking mode reasoning provide the most significant contributions.
The framework demonstrates that structured search effectively complements neural language models for contamination-free benchmarks, providing a foundation for future work on multi-file modifications and integration with static analysis tools.

\end{document}